\documentclass[journal]{IEEEtran}

\ifCLASSINFOpdf
 
\else
 
\fi

\usepackage{amsmath}

\usepackage{graphicx}
\usepackage{caption}
\usepackage{subcaption}
\usepackage{booktabs}

\begin{document}

\title{FViT-Grasp: Grasping Objects With Using Fast Vision Transformers}

\author{
    \IEEEauthorblockN{Arda~Sarp Yenicesu\IEEEauthorrefmark{1}\IEEEauthorrefmark{2}, Berk~Cicek\IEEEauthorrefmark{1}\IEEEauthorrefmark{2}, Ozgur S. Oguz\IEEEauthorrefmark{2}}
    \thanks{\IEEEauthorrefmark{1}Equal Contribution}
    \thanks{\IEEEauthorrefmark{2}Computer Eng. Dept., Bilkent University., Türkiye}
}



%



\maketitle

\begin{abstract}
This study addresses the challenge of manipulation, a prominent issue in robotics. We have devised a novel methodology for swiftly and precisely identifying the optimal grasp point for a robot to manipulate an object. Our approach leverages a Fast Vision Transformer (FViT), a type of neural network designed for processing visual data and predicting the most suitable grasp location. Demonstrating state-of-the-art performance in terms of speed while maintaining a high level of accuracy, our method holds promise for potential deployment in real-time robotic grasping applications. We believe that this study provides a baseline for future research in vision-based robotic grasp applications. Its high speed and accuracy bring researchers closer to real-life applications.
\end{abstract}

\begin{IEEEkeywords}
robotic, manipulation, grasping, pick-and-place, deep-perception
\end{IEEEkeywords}

\IEEEpeerreviewmaketitle

\section{Introduction}

\IEEEPARstart{G}{rasping} is an essential ability for robots to interact with and manipulate objects in their environment. While humans can intuitively grasp objects, this is a challenging task for robots, requiring the ability to perceive and understand their surroundings. In the past, researchers tried using camera geometry and 3D models of objects to enable robots to grasp objects, but these approaches are limited in practicality and real-world applications. Deep visual perception, which uses deep learning to analyze visual data and make decisions, has shown potential in enabling robots to grasp a wide range of objects in unstructured environments. This approach allows robots to infer their surroundings and find a suitable grasping pose without needing full state knowledge or human-crafted features.

In this research, we propose a novel method to enable robots to detect suitable grasping poses for interacting with objects in their environment. Instead of employing a convolutional neural network (CNN) as a backbone for grasp prediction, we utilize a vision transformer known as LITv2 [2]. This approach involves using LITv2 to predict a global grasp configuration for an input scene through regression, rather than classification [3].

We observed that this approach achieved a remarkable accuracy of 88.2[\%] while operating at a rapid speed of 150 frames per second. These results are particularly noteworthy due to the importance of the backbone architecture's speed in applications within the robotics domain, where the model needs to function online. Our evaluation utilized the Cornell Grasp Dataset [6], sourced from a Kaggle dataset [7] as the official version is no longer available. Our findings indicate that our approach surpasses existing solutions in terms of speed while maintaining a high level of accuracy.

We integrated our self-developed model with the Drake environment to simulate and test our results in a robotics experiment. Initially, we created a scene in Drake comprising two bins, each containing an object from the YCB dataset, along with a 7-DOF Kuka iiwa arm. In our experiment, we devised a task for the robot to pick up and transfer an object from one bin to another. The input RGB image is captured by a camera in the environment, processed through our deep learning model, and predicts possible grasp locations, which are indicated with bounding box coordinates as an output. The robot then focuses only on the indicated area to find an appropriate grasping point. This process not only saves computational time but also enhances efficiency by avoiding a search through the entire area.

In Section IV, we provide an explanation of our proposed model, and Section V is dedicated to presenting the results of our study.

\section{Related Work}
There are four main categories of deep visual perception-based approaches for robotic grasp detection: regression-based, classification-based, generative, and region proposal-based approaches. Regression-based approaches involve using a convolutional neural network (CNN) and fully connected layers as a regression head to predict a single global grip across all input images. One example of this is the work of [8], which used the pre-trained AlexNet [1] CNN as a backbone and fine-tuned it with their own regression head. [9] also used a regression-based approach but modified the ResNet [4] architecture and used two fully connected layers at the end of ResNet-50 as their regression head.

Classification-based approaches involve using a CNN and fully connected layers as a classification head to predict the class of a grasp configuration. Generative approaches involve using a CNN and fully connected layers to generate a grasp configuration.

Region proposal-based approaches use a CNN and fully connected layers to suggest regions that may contain grasp configurations. These regions are then refined through additional processing. [10] proposed a constrained approach to robotic grasp detection that involves classifying the orientation of grasping rectangles with a fixed size using a CNN with 18 bins to discretize the orientation. However, this approach may not be generalizable because it assumes a fixed size for the grasping rectangles. [11] used a fully-convolutional network to predict the grasp configuration for each pixel in the input scene, outputting 3 images with the same resolution as the input that represent the quality, width, and angle of the grasp configuration at each pixel. This approach allows for multiple grasp configurations to be generated for an input scene, which is particularly useful in cluttered environments. [12] improved upon this approach by using a modified version of the ResNet architecture to increase the depth of the network, resulting in improved performance.

[13] combined the advantages of regression-based and generative approaches by introducing a Grasp Proposal Network, which is a robotic adaptation of the region proposal networks (RPNs) used in object detection. This network performs multiple grasp rectangle regressions in candidate regions using the ResNet-50 as the backbone. [14] improved upon this approach by using semantic segmentation to isolate the target object, resulting in more accurate grasp configuration estimates in cluttered scenes.

Another common approach for grasping action is using a geometric model of the object and matching the object model and scene model using ICP. In [15], authors use the scheme of getting input from RGB-D image and segment the objects. After that, 3D model pose estimation and evaluation are done. For the pose estimation, ICP is used, and in ICP, the initial guess of the points is important. In our method, we also try to supply a good initialization. The strongest point of geometric model matching according to other methods with DL is that it does not require training. Some studies try to find model-free methods for object pose estimation [20]. This method offers pose estimation without requiring a 3D CAD model of the object. These methods can also be combined for end-to-end grasp point detection for objects in future studies.

\section{Dataset}
\begin{figure}[htbp]
\centerline{\fbox{\includegraphics[scale=0.50]{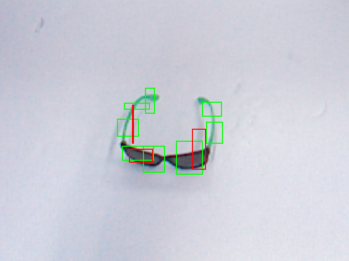}}}
\caption{The green bounding box in this example from the Cornell Grasping Dataset indicates a positive grab, whereas the red bounding box indicates a negative grasp.}
\label{cgd2}
\end{figure}

The Cornell Grasp Dataset consists of images of a single object scene with annotations on successful and unsuccessful grasps. It includes a total of 885 images of 240 objects, with 5110 positive and 2909 negative grasp examples labeled by humans. These labels are not necessarily exhaustive, and the number of labels per grasp can vary. When training machine learning models using this dataset, a single random positive label is selected, while all labels are used during the testing phase. The dataset can be used to train models to predict successful grasps. An example of the grasp rectangles in the dataset is shown in Fig.~\ref{cgd2}.

\section{Representation}
\begin{figure}[ht]
\centerline{\includegraphics[scale=0.22]{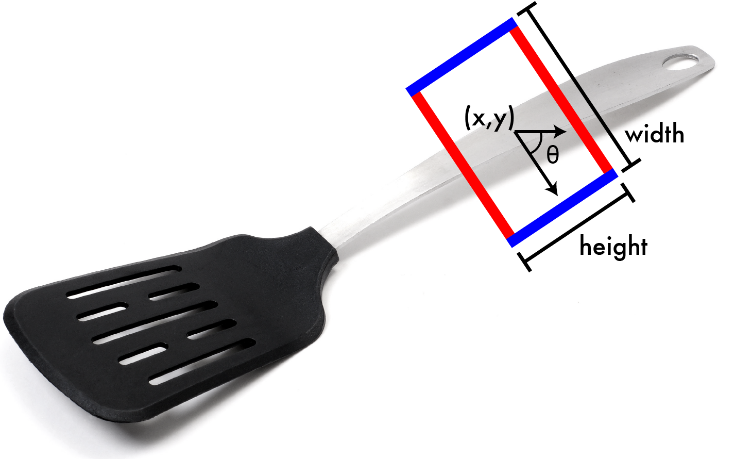}}
\caption{5 dimensional representation}
\label{5dim}
\end{figure}

To represent a grasping rectangle on an RGB image, two widely adapted notions exist. The first one is the 8-dimensional representation, where each corner of the rectangle is represented by two values: x and y. The second representation is called the 5-dimensional representation. It is adapted from object detection terminology in which the rectangle is represented by the center, width, height, and the horizontal rotation with respect to the center
\begin{equation}
    Grasp Rectangle = (x, y, \theta, w, h)
\end{equation}
Fig.~\ref{5dim} shows the 5 dimensional representation. 

\section{Metric}
To compare our model with the state-of-the-art and to evaluate the proposed grasping rectangles, we used the rectangle metric, which determines the accuracy of the proposed grasping rectangle. It decides on accuracy based on two conditions. Grasping is determined as accurate if:
\begin{itemize}
    \item Jaccard similarity index should be higher than 25\% between the rectangles.
    \begin{equation}
        Jaccard(U,V) = \frac{|U \cap V|}{|U \cup V|}
    \end{equation}
\end{itemize}

\section{Method}We introduce a novel application of a deep learning architecture, FViTGrasp, which can infer the target grasp region of an object in a single scene scenario without having the full state knowledge. Since we want to perform a robust grasping operation, we further combine this inference with a geometric approach, where we filter the 3D point cloud using the proposed grasp region and find the antipodal grasping point by sampling the filtered point clouds. We aim to increase the efficiency and performance of this geometric grasp approach by using the observations obtained from our model's deep perception. In this section, we will discuss our architecture, how to transfer deep perception output to the simulation environment to filter the point cloud data, and how to generate antipodal grasp candidates.
\subsection{Architecture}
\begin{figure}[ht]
\centerline{\includegraphics[scale=0.33]{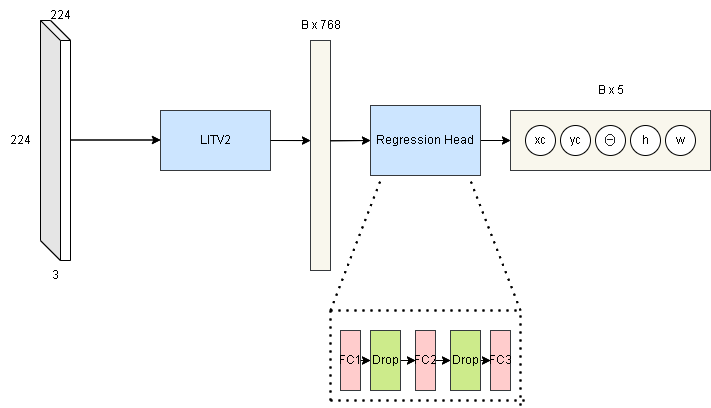}}
\caption{General structure of our model.}
\label{model}
\end{figure}

The fast vision transformer with HiLo attention mechanism, LITv2[2], is intended to take the role of the convolutional neural network backbone in the proposed design, FViTGrasp. We set out to build a model that is faster than the state-of-the-art without significantly sacrificing accuracy across the whole design process. To do this, we used the global grasp regression method, in which we forecast the ideal grasp configuration from a scene of a single item. An overview of our architecture, which we employed to address the robotic grasp detection problem, is shown in Fig.~\ref{model}.

We further describe the backbone and the regression head in the following subsections.

\subsubsection{Backbone}
\begin{figure}[ht]
\centerline{\includegraphics[scale=0.13]{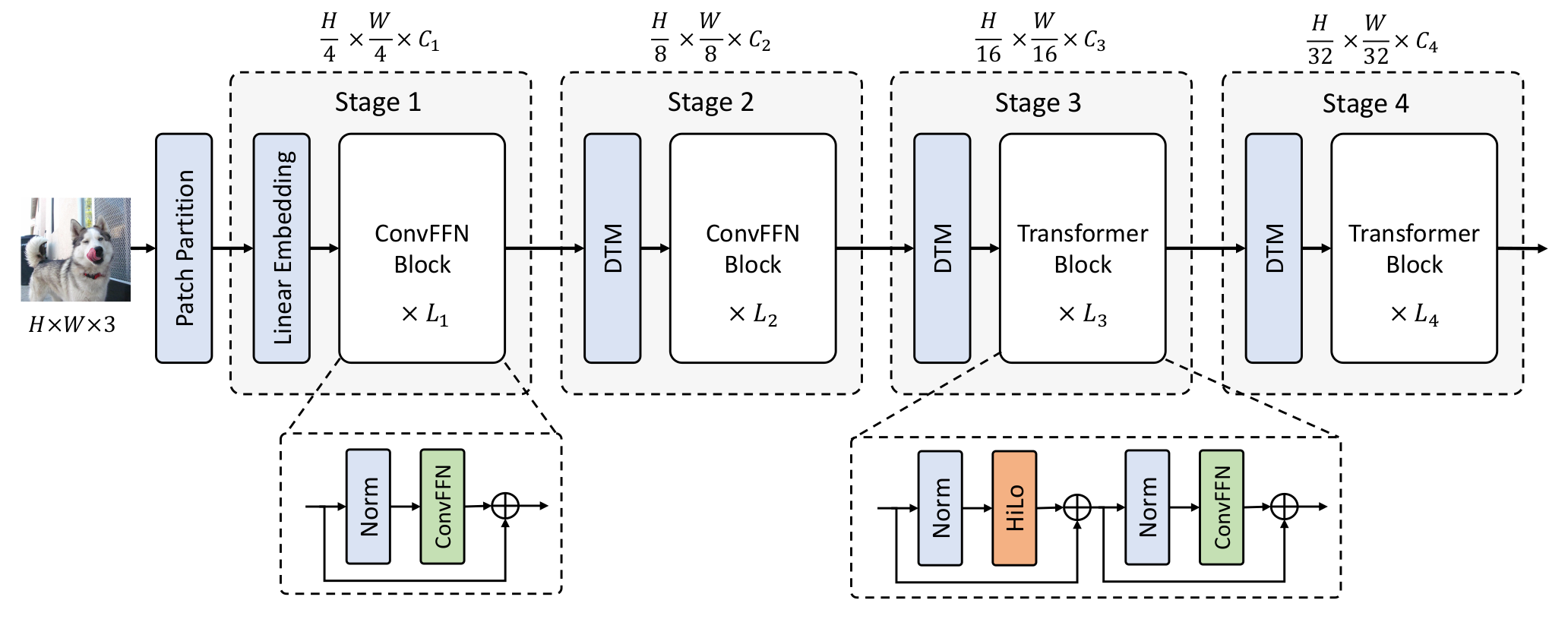}}
\caption{General structure of the LITv2 [2].}
\label{LITv2}
\end{figure}

In order to focus more on speed than accuracy, we choose to employ the LITv2 small version [2], which performs at a throughput of 1471 frames per second while suffering a loss of 2.7 top-1 accuracy in comparison to its deepest version in the image classification challenge on ImageNet-1K. In order to develop a model that is more suitable with real-world settings, we accept this possible trade-off in our application. The overall architecture of the suggested vision transformer in [2] is seen in Fig.~\ref{LITv2}.

\begin{figure}[ht]
\centerline{\includegraphics[scale=0.11]{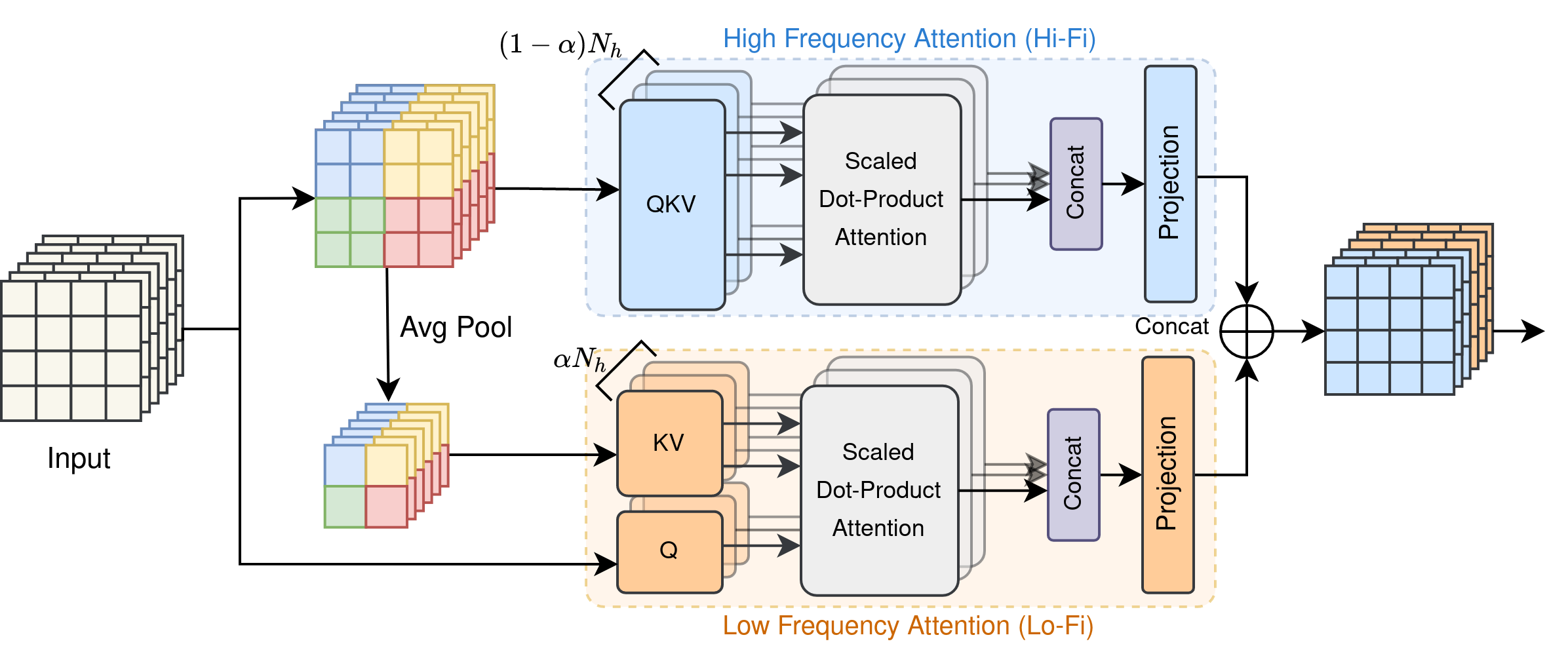}}
\caption{HiLo attention block [2].}
\label{hilo}
\end{figure}

The effectiveness of vision transformers is constrained in small data regimes, as in our case because the Cornell Grasp Dataset is a relatively small one, since its inductive bias is lower compared to the Convolutional Neural Networks. Because of this, we applied additional fine-tuning to the LITv2-S version that had already been trained for classification tasks using the ImageNet-1K dataset. This backbone proposes a novel attention mechanism, HiLo attention, to increase its efficiency and performance, motivated by the idea that in natural images high frequencies capture the fine details while low frequencies cover the global structure. With this separation, high frequency heads use windowed attention and low frequency heads use average pooling, both of which increase the speed compared to regular multi-head self-attention. Fig.~\ref{hilo} shows the proposed HiLo attention block [2]
\\
\subsubsection{Regression Head}
We flattened the last layer of the LITv2-S to a 768-dimensional representation and utilized three fully connected layers with leaky Relu activation and dropout layers in between the fully connected layers as our regression head in order to predict the optimal grip configuration. The following table displays the general design of our regression head.
\begin{table}[ht]
\renewcommand{\arraystretch}{1.3}
\caption{Regression Head}
\label{table_regression}
\centering
\begin{tabular}{|c|}
\hline
Layers\\
\hline
FC1: (768, 2048)\\
\hline
Leaky Relu with 0.1 leak\\
\hline
Dropout with 0.1 probability\\
\hline
FC2: (2048, 1024)\\
\hline
Leaky Relu with 0.1 leak\\
\hline
Dropout with 0.1 probability\\
\hline
FC3: (1024, 5)\\
\hline
\end{tabular}
\end{table}\\
The output is a 5 dimensional grasp configuration but for the version that we have used in the Drake, we decided to output 8 dimensional polygonal representation for the target grasp region since it will be a smoother representation of the binary point cloud filter.
\subsection{Camera Frame to World Frame Transformation}
To achieve end-to-end simulation, transferring the points in camera frame to world frame is a crucial step. Since we get the bounding box in image plane, we have to transform it to world frame to detect the position of the object in the simulated world. One of the most conventional methods to achieve this transformation is using pinhole camera model. 
\begin{figure}[ht]
\centerline{\includegraphics[scale=0.49]{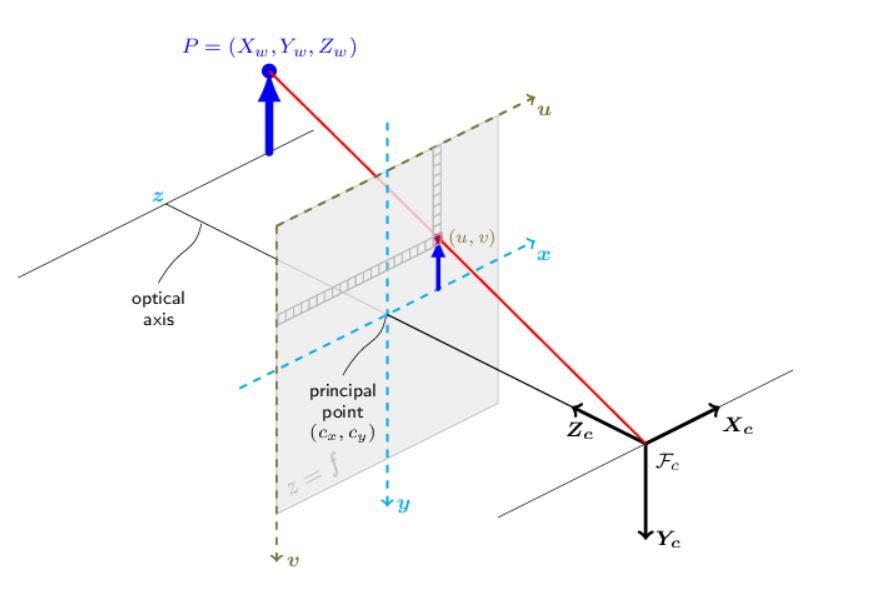}}
\caption{Pinhole Camera Model [18].}
\label{pin}
\end{figure}\\
\begin{equation}
    X_{c} = (u-c_{x})\frac{Z_{c}}{f_{x}}
\end{equation}
\begin{equation}
    Y_{c} = (v-c_{y})\frac{Z_{c}}{f_{y}}
\end{equation}
These equations above provide the transformation from the image plane to the simulation world. Here, $f_{x}$ and $f_{y}$ represent the focal lengths in the x and y directions, while $c_{x}$ and $c_{y}$ denote the principal points in the x and y directions, respectively. Additionally, $X_{c}$, $Y_{c}$, and $Z_{c}$ signify points in the camera frame.

A critical aspect is determining the Z-coordinate of the object in the camera frame. As we are using an RGB image, there is no inherent information available regarding the Z position of the object. Consequently, we initialize a Z coordinate as a hyperparameter. This coordinate is defined while taking into account the physical constraints. Given that objects cannot exist outside the bin, we establish the lower and upper layers of the bin (box) as the boundaries. Thus, the Z coordinate is constrained within this range. These coordinates delineate an area, within which we retain the point clouds, discarding those lying outside.

Subsequently, after computing the cropped point cloud of the object in the world frame, the subsequent step involves generating antipodal points as potential candidates.
\subsection{Generating Antipodal Grasping Candidates}
The points within the point cloud undergo cropping using the projection method outlined in the preceding section. This results in a downsampled version of the point cloud, containing only the points relevant for the grasping task. From this downsampled set, we randomly select a point.

It's worth noting that our downsampling approach enhances the efficiency of the sampling process, as we are working with a subset of the target points. To obtain the local geometry of the sampled point, we estimate its normal vector by analyzing a patch in its vicinity, as follows:

\begin{align*}
    \intertext{\centering
        $p^i$  : ith point in the point cloud \\
        $p^*$  : mean of the point cloud \\
        $N$    : Number of points in a patch \\
        $n$    : Normal vector corresponding to the sampled point
    }
\end{align*}

\begin{equation}
    p^* = \frac{1}{N}\sum_{n=1}^{N} p^i \quad \textrm{N: Number of points in a patch}
\end{equation}
\\
\begin{equation}
    W = \left[ \sum_i  (p^i- p^*) (p^i - p^*)^T \right]
\end{equation}
\\
\begin{equation}
    \min_n n^T W n, \quad \text{s.t} \quad |n|=1 \quad 
\end{equation}
\\
The eigenvector corresponding to the smallest eigenvalue of matrix W provides the optimal normal vector based on the local patch. Once we have determined the normal vector corresponding to the sampled point, we align it with the x-axis of our two-finger gripper. We then sample orientations relative to this normal. Ideally, this point should have an antipodal pair that makes contact with the other finger of the gripper.
\begin{figure*}[!htb]
    \centering{}
    \begin{subfigure}{0.2\linewidth}
        \centering{}
        \fbox{\includegraphics[width=\linewidth, height=0.3\textheight, keepaspectratio]{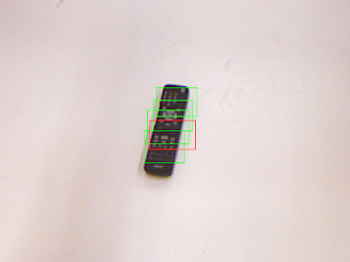}}
        \fbox{\includegraphics[width=\linewidth, height=0.3\textheight, keepaspectratio]{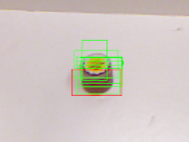}}
        \caption{True positives.} \label{good}
    \end{subfigure} 
    \begin{subfigure}{0.2\linewidth}
        \centering{}
        \fbox{\includegraphics[width=\linewidth, height=0.3\textheight, keepaspectratio]{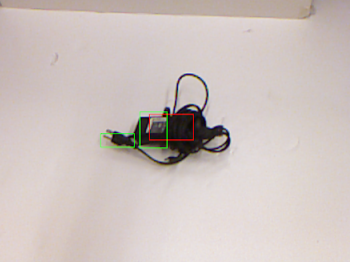}}
        \fbox{\includegraphics[width=\linewidth, height=0.3\textheight, keepaspectratio]{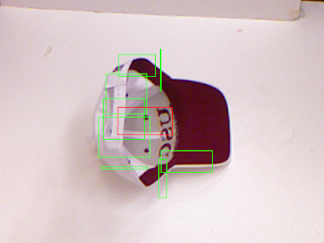}}
        \caption{True negatives.} \label{bad}
    \end{subfigure} 
    \begin{subfigure}{0.2\linewidth}
        \centering{}
        \fbox{\includegraphics[width=\linewidth, height=0.3\textheight, keepaspectratio]{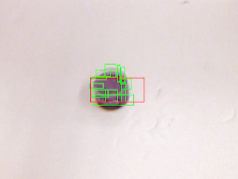}}
        \fbox{\includegraphics[width=\linewidth, height=0.3\textheight, keepaspectratio]{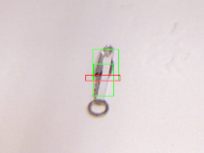}}
        \caption{False negatives.} \label{badgood}
    \end{subfigure} 
    \caption{Grasp examples. (a) True positives define the accurate grasp.(b) True negatives define the unsuccessful grasp. (c) False negatives defines the failure cases but potential grasping in real-world scenarios. The green rectangles represent the ground truth grasp rectangles, while the red rectangles represent our model's prediction.}
    \label{graspex}
\end{figure*}\\
To assess these orientations, we rank the grasps based on their antipodal scores, considering factors such as:
\begin{itemize}
    \item Are the normal vectors produced by gripper finger's contact antipodal enough?
    \item Does the orientation violate the non-penetration constraint?
\end{itemize}
We select the highest scoring gripper pose as our antipodal grasping candidate.

\subsection{Grasping of an Object}
After obtaining the point cloud for picking up the object, the subsequent step is to search for valid antipodal points suitable for grasping. The primary objective is aligning the robot's finger with normal points randomly selected from the downsampled point cloud. Costs are then computed, and the point with the lowest cost is chosen as the optimal grasp point.

Once the points are identified, the planner takes over to handle the task and motion planning. Initially, the planner determines the task based on the simulation's state. For instance, if there is no object in the first bin, it proceeds to check the second bin. If an object is found in the second bin, it moves on to the next task, which involves transporting the object to another bin. Grasping is planned by the motion planner, which employs inverse kinematics.

\section{Experiments \& Results}
We used the NVIDIA GeForce RTX 3070 Mobile GPU for all of our experiments, and PyTorch and Drake framework to construct our approach. First we wanted to try our approach on well-adapted baseline dataset, Cornell Grasp Dataset[6]. A table II compares our model and other methods for grasping accuracy and speed in terms of frames per second.
\begin{table}[ht]
\renewcommand{\arraystretch}{1.3}
\caption{Grasping Accuracy and Speed Results}
\label{table_example}
\centering
\begin{tabular}{lccccl}\toprule
Method & Input &  Grasp Acc. (\%) & Speed (FPS)\\\midrule
Zhou [16], Resnet-50 & RGB & 97.7 & 9.9\\
Zhou [16], Resnet-101 & RGB & 97.7 & 8.5\\
Zhang [17], Resnet-101 & RGB & 93.6 & 25.2\\
Karaoguz [19], VGG-16 & RGB & 88.7 & 2\\
Chu [13], Resnet-50 & RGB & 94.4 & 8.3\\
Kumra [12], GR-ConvNet & RGB & 96.6 & 52\\
Det\_Seg\_Refine [14] & RGB & \textbf{98.2} & 63\\
\textbf{FViT-Grasp (ours)} & RGB & 88.2 & \textbf{150}\\\bottomrule
\end{tabular}
\end{table}\\
With an average of 150 frames per second, we have exceeded the previous state-of-the-art in terms of speed (FPS) for robotic grip detection research. Additionally, we have improved upon the previous study [8] by 3.8\%, achieving a new state-of-the-art among all robotic grasp detection studies that use global grasp prediction, achieving an accuracy of 88.2\%. The following outputs display some examples of successful, unsuccessful, and feasible predictions made by our model on samples from the Cornell Grasp Dataset[6].

Fig.~\ref{good} shows accurate grasp predictions. One possible comment is that our model is accurate when the scene has a clear foreground-background separation.

Fig.~\ref{bad} shows inaccurate grasp predictions. For the first one, cable makes the object non-rigid, so that it is harder for the model to generalize an accurate grasp for this object. In the second one, it is clear to observe that there is no distinct separation between background and the foreground, which forces our model to fail. Lastly, a third failure case shows an object with a changing depth which may affect the grasping strategy, but our model only relies on the RGB image so that it fails in such scenarios.

Lastly, Fig.~\ref{badgood} displays grasps that are considered failures according to the rectangle metric, but they might be feasible when applied in a real-world scenario. In the first example, both the ground truth and predicted grasps change in size, but they both propose a valid grasping. The second one falls outside the rotation range of the given labels, yet it still represents a valid grasping proposal.

Our model excels at identifying grasping configurations when there's a clear distinction between the object and the background. However, it struggles in scenarios where additional information is required, such as in the example with the hat where the depth of the object changes throughout the scene. This has motivated us to explore geometrical approaches for grasping.\\

\begin{figure*}[!htb]
    \centering{}
    \begin{subfigure}{0.2\linewidth}
        \centering{}
        \includegraphics[scale=0.40]{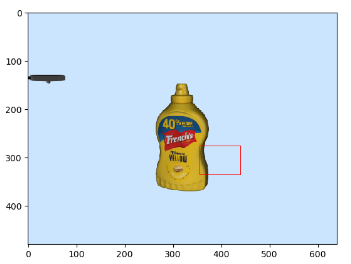}
        \caption{Without refinement.} \label{drakebad}
    \end{subfigure}
    \hspace{2cm} 
    \begin{subfigure}{0.2\linewidth}
        \centering{}
        \includegraphics[scale=0.35]{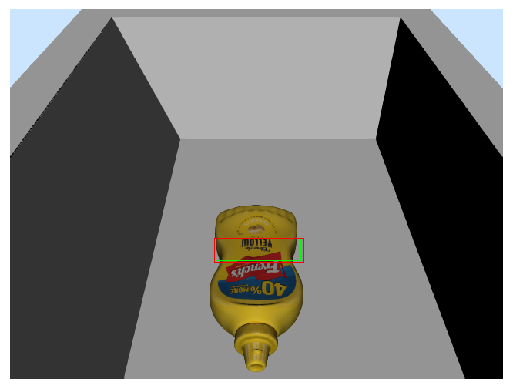}
        \caption{With refinement.} \label{drakegood}
    \end{subfigure} 
    \hspace{2cm} 
    \begin{subfigure}{0.2\linewidth}
        \centering{}
        \includegraphics[scale=0.35]{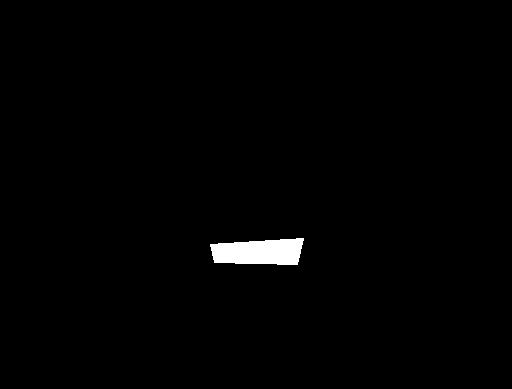}
        \caption{Polygon mask.} \label{mask}
    \end{subfigure} 
    \caption{(a) Grasp prediction without refinement on Drake. (b) Grasp prediction with refinement on Drake. (c) Binary polygon mask that highlights the grasp target region.}
    \label{DD2}
\end{figure*}

Once we confirmed the success of our approach in finding grasps using RGB images, we proceeded with our robotic experiments using the Drake Simulation. Fig.~\ref{drakebad} showcases one of the best examples of grasping if we were to apply our model directly in the Drake environment.

The potential causes of this inaccurate grasp prediction in the simulation environment are the differences in the camera perspective and background differences. However, after our extensive experiments on different backgrounds, objects and rotations, we are confident to say that the model that is trained on Cornell Grasp Dataset cannot perform well, at least with our augmentations. For this reason, we construct a new, small data set containing only scenes from the Drake simulation environment and trained our model from scratch to be able to perform in the Drake environment. Fig.~\ref{drakegood} shows an output of the model adjusted for the drake environment.

The output of the adjusted model provides an accurate grasping according to the rectangle metric. However, due to the representation purposes it seems like a perfect rectangle, but it is actually a polygon. As we mentioned before, we want to combine geometric approaches with the deep perception. For this purpose, we treat this rectangle as possible grasp region so that we can use it to crop the point cloud and perform a robust grasping in the simulation environment with the methods we mentioned before. Fig.~\ref{mask} shows the polygon binary mask that we will use to filter point clouds.

\begin{figure*}[t]
    \centering
    \begin{subfigure}[b]{0.6\linewidth}
        \centering
        \includegraphics[width=\linewidth, height=0.3\textheight, keepaspectratio]{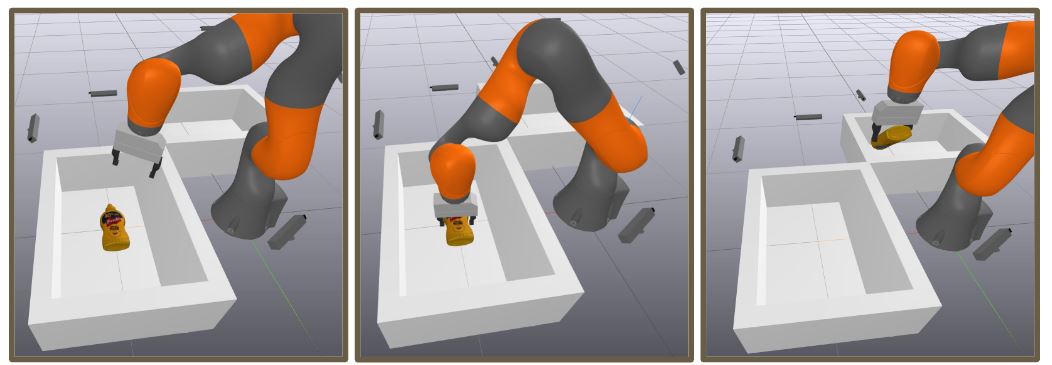}
        \caption{Straight object.} \label{simulation}
    \end{subfigure} \\[10pt]

    \begin{subfigure}[b]{0.6\linewidth}
        \centering
        \includegraphics[width=\linewidth, height=0.3\textheight, keepaspectratio]{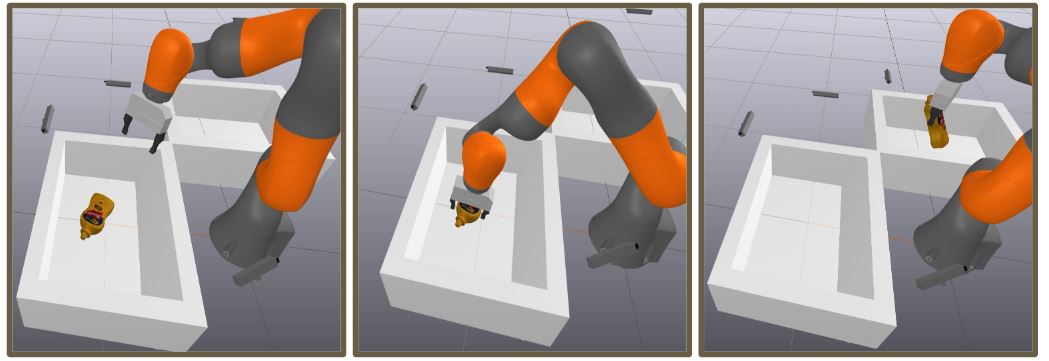}
        \caption{Rotated object.} \label{simulation2}
    \end{subfigure} 
    \caption{Before grasp, during grasping and after grasping poses of simulation.}
    \label{graspex}
\end{figure*}

After obtaining the mask, coordinates of corner points are selected and the point clouds outside the selected area are eliminated from the grasping points search space. This region is transformed from image plane to world frame with using the homography which we also discussed in Section IV.B.
After this, the point cloud in that region is cropped and transferred as an input to the antipodal point calculator function to achieve valid grasping points. Then, with motion planner, the object is grasped and transformed from one bin to another. In the Fig.~\ref{simulation} three states of the simulation can be seen. In the first image, the state of the robot arm is represented just before the grasping action. The second image represents the robot during grasping action. The last image shows the state of the robot and the object while the object is carried to the second bin.

In the Fig.~\ref{simulation2} three states of the simulation with different initial orientation of the object can be seen. Again, in the first image, the state of the robot arm is represented just before the grasping action. The second image represents the robot during grasping action. The last image shows the state of the robot and the object while the object is carried to the second bin. These results presented as an example of that our model can also work with different orientation of the object.\\
In addition, we also investigate the contribution of our model in terms of speed when it is compared to geometrical approaches. For this purpose, we ran ten grasping trial both with solely geometrical approaches and hybrid approach with deep perception (ours), and average the speed of these trials. Speed is calculated in terms of the time passing from starting of simulation to the  completion of motion planning. With using geometrical approach, average time takes to detect grasping object and planning the motion is 5.04 second. This time for our method is 4.38 second. These quantitative results show that our model improves the speed around 15$\%$. For a simple case this quantity can be less conspicuous, but for more complex and long cases this improvement has large impact.

\section{Discussion}
Our approach was successful to perform an accurate and efficient grasp by combining the deep visual perception and geometric grasping approaches; however, we would like to point out some insights that might be useful for the future studies.\par
As we mentioned before, our original model which is trained on real RGB images from the Cornell Grasp Dataset did not perform well at Drake's simulation environment as we couldn't produce reasonable grasp configurations with it and had to train our model from scratch with synthetic samples. Our first assumption about this failure was different illuminations, different camera perspective, out of distribution target object. However, after extensive investigation we believe that main complication is the texture details and characteristics are very different in pixel domain for model to generalize the scenes between real and drake environments, even if they are close to each other from human's perspective.\par
Additionally, while training with the generated samples from the Drake environment, training with randomly initialized weights did perform better than the one with the pre-trained weights at an image classification on ImageNet-1K dataset. This was an interesting experimental outcome since pre-training is a common practice used to train visual transformers, which suggest that texture information for the RGB scenes differs from the ones we have obtained in the simulation. One can claim that transfer learning is inapplicable from real scenes to Drake environment, even though inverse is common in general robotic applications.\par
As a future work, we want to generate more grasping samples with trial-and-error using the geometric approaches to cover wider distribution and increase the performance of our model in drake. Moreover, we will investigate the possible ways to increase the deep perception's contribution to have better generalizability in our hybrid grasping approach.
\section{Conclusion}
In this study, we introduced a novel architectural solution, FViT-Grasp, leveraging HiLo attention and a fast vision transformer, to address the critical challenge of robotic grip detection. Through rigorous experimentation in a simulated robotic environment using Drake, we employed a hybrid approach that synergizes deep visual perception with geometric grasping techniques. Remarkably, our model demonstrated an impressive operational speed of 150 frames per second, surpassing previous benchmarks, while concurrently elevating the accuracy of models utilizing the global grasp prediction method within the Cornell Grasp Dataset.

One of the significant findings of our study pertains to the behavior of transfer learning between real-world and simulated scenes. Notably, we observed that training our model exclusively on simulated data yielded superior results within the Drake environment. This crucial insight sheds light on the dynamics of transfer learning in the context of real versus simulated environments.

Our work represents a substantial step forward in the field of robotic grasp detection, offering an innovative approach that balances speed and accuracy. Nevertheless, we acknowledge that challenges persist, particularly in bridging the gap between real-world and simulated data. To this end, future research endeavors will focus on generating more diverse grasping samples through trial-and-error using geometric methodologies, aiming to broaden the model's capability and improve performance within the Drake simulation.

In conclusion, our study not only presents an architectural solution but also provides valuable insights into the complexities of transfer learning in simulated environments. We anticipate that our work will inspire further advancements in the field of robotic grasp detection, ultimately paving the way for more effective and efficient robotic interactions with the physical world.

\ifCLASSOPTIONcaptionsoff
  \newpage
\fi

\end{document}